# Evaluating GPT's Programming Capability through CodeWars' Katas

Zizhuo Zhang, Lian Wen, Shaoyang Zhang, David Chen , Yanfei Jiang
Chang'an University, Xi'an, China, zzha4913@chd.edu.cn
Griffith University, Brisbane, Australia, l.wen@griffith.edu.au
Chang'an University, Xi'an, China, zhsy@chd.edu.cn
Griffith University, Brisbane, Australia, david.chen@griffith.edu.au
Xi'an Rail Transit Group Company Limited, Xi'an, China,  flyjyf@sina.com

*Abstract*— In the burgeoning field of artificial intelligence (AI), understanding the capabilities and limitations of programming-oriented models is crucial. This paper presents a novel evaluation of the programming proficiency of Generative Pretrained Transformer (GPT) models, specifically GPT-3.5 and GPT-4, against coding problems of varying difficulty levels drawn from Codewars. The experiments reveal a distinct boundary at the 3kyu level, beyond which these GPT models struggle to provide solutions. These findings led to the proposal of a measure for coding problem complexity that incorporates both problem difficulty and the time required for solution. The research emphasizes the need for validation and creative thinking capabilities in AI models to better emulate human problem-solving techniques. Future work aims to refine this proposed complexity measure, enhance AI models with these suggested capabilities, and develop an objective measure for programming problem difficulty. The results of this research offer invaluable insights for improving AI programming capabilities and advancing the frontier of AI problem-solving abilities.

## I. INTRODUCTION

As we continue to navigate the digital age, artificial intelligence (AI) capabilities are rapidly advancing, significantly impacting various societal facets. One of these remarkable advancements is the development of natural language processing (NLP) models, a field exemplified by the Generative Pretrained Transformer (GPT) series developed by OpenAI [1].

Several studies have been conducted to evaluate different aspects of GPT's programming capability. These investigations include testing ChatGPT for numerical methods [2], utilizing ChatGPT to find programming bugs [3], refactoring and generating test cases [4], and detecting code vulnerabilities [5]. Previous research indicates that while GPT might struggle with answering multiple-choice questions about code [6], it can assist students in learning programming [7], though it's not yet capable of directly passing university programming course examinations. GPT can help make code suggestions but has only shown 57% correctness in Python and 27% correctness in Javascript when solving Leetcode problems [8] [9]. Karmakar's test of 115 Python problems from HackerRank [10] on Codex indicated that it was able to solve 96% of the problems [11]. However, there are signs that Codex may generate memorized code as its solution.

Despite these studies, no research has yet attempted to evaluate GPT's performance on coding questions at various difficulty levels. This paper offers an innovative perspective, evaluating the programming prowess of GPT models in this aspect.

Our experiment involved GPT-3.5 and GPT-4, as provided on OpenAI's website [1]. We tested them on eight different difficulty levels of problems from Codewars [12], a website that offers coding exercises, known as 'katas.' Each kata has an associated difficulty level, ranging from 8kyu (easiest) to 1kyu (hardest), based on numerous programmers' reviews [13]. Our findings indicate that GPT-4 is more reliable and can solve all questions from 4kyu to 8kyu when given feedback. However, we observed a clear boundary at the 3kyu level that posed a significant challenge to all tested GPT models, leaving questions from 1kyu to 3kyu unsolved.

In human problem-solving, we identify three distinct types of intelligence: intuition, validation [14], and creative thinking [15]. The 'intuition' intelligence, where the mental process automatically generates output from input, can be emulated in AI through neural networks [16]. Despite the extensive training times, these models can respond swiftly with minimal delay, with the quality of the translation significantly dependent on the quality of the training.

'Validation' intelligence is required, especially for coding problems, to verify solutions against a comprehensive set of test cases and to correct bugs in the original output. Current GPT models may lack this capability, often requiring human intervention to determine the correctness of their solutions.

Lastly, the most challenging type of intelligence is 'creative thinking.' When faced with a complex problem, humans may ponder for a considerable time, seemingly make no progress, and then suddenly conceive a solution, either with or without noticeable external inspiration.

Based on these observations, we propose that for AI to better emulate human intelligence, it needs to acquire both validation and creative thinking capabilities. The existence of this boundary in GPT's problem-solving abilities and its implications are largely due to the design of the AI models.

These insights, along with our proposed measure of coding problem complexity, form the major contributions of this paper. They offer a deeper understanding of both the potential and limitations of AI in programming.

The novelty of this research lies in evaluating ChatGPT's programming capability based on different levels of questions. A significant discovery is the existence of a clear difficulty boundary for GPTs. We speculate that this boundary is a result of the design of the current AI models for GPT. To break this boundary and reach the highest human programming level, we propose the addition of efficient validation models and creative models. This research also implies a need to design an objective measure of a programming problem's difficulty level.

The subsequent sections delve into the background of the GPT model and Codewars' Katas (Section 2), detail our experiment (Section 3), provide a comprehensive discussion of the experiment's results (Section 4), and conclude with prospects for future research (Section 5).

## II. BACKGROUND

ChatGPT [1], a generative pre-trained transformer model, has made significant strides in natural language processing and understanding. Despite these advances, there are several challenges that must be addressed to effectively solve programming problems.

In this section, we first discuss challenges to use AI to solve programming problems and then explain how ChatGPT addresses these challenges.

After that we briefly introduce Codewars, and explain why we select Codewars problems as a way to evaluate ChatGPT's programming level.

*2.1 Challenges for AI to Solve Programming Problems*

In recent decades, AI has made impressive progress in various areas and even exceed human's intelligence in certain tasks, for example, play board games. Solving programming problems presents several challenges for AI.

Firstly, the ambiguity in problem statements can be difficult for AI to interpret. Unlike humans who can often infer missing details, AI lacks this intuitive understanding [17].

Secondly, programming problems often require a deep understanding of the problem domain and relevant context, which is hard for AI to acquire. AI models like GPT-3 or Codex, while skilled at generating code, can still produce incorrect or nonsensical outputs due to lack of context [18] [9].

Thirdly, the creative aspect of programming is a challenge for AI. Creative problem solving: Unlike traditional programming problems, where the number of valid solutions is often limited, real-world programming tasks may require creative thinking and the ability to devise novel solutions [19]. While AI can reproduce patterns, it can struggle with creating innovative solutions, which is often needed in programming [20].

Lastly, verifying the correctness of a generated solution is a challenge. It requires AI to not only generate the code but also verify its correctness against all possible edge cases, which is computationally challenging [21]. These challenges necessitate continuous development and training of AI models to better understand, generate, and verify code.

*2.2 How Large Language Models Address those Challenges*

Based on literature, this paper discusses how ChatGPT addresses these challenges and acquire recognizable programming capability.

1. *Understanding problem context*: ChatGPT, as a language model, is trained on vast text corpora that include various programming-related materials, enabling it to understand programming languages, concepts, and problem statements to some extent [22]. However, fully comprehending the context of programming problems remains a challenge. Recent work on few-shot learning [18] and incorporating specialized knowledge [23] aims to improve ChatGPT's ability to understand problem context and generate more accurate code snippets. While progress has been made, further research and model refinements are necessary to achieve a deeper understanding of programming problem contexts.

2. *Code validation and debugging*: ChatGPT's current limitations in code validation and debugging can be attributed to its primary focus on natural language understanding. To address this challenge, researchers have proposed integrating specialized components into GPT models to enable self-validation and debugging capabilities [24]. This could involve incorporating techniques from program synthesis, static analysis, and runtime analysis to iteratively refine generated code [25]. Although preliminary research has explored these directions, the integration of code validation and debugging capabilities into ChatGPT remains an ongoing challenge.

3. *Creative problem solving*: Encouraging creative problem-solving in ChatGPT requires the development of techniques that can push the model to generate more diverse and innovative solutions. Recent work on models like DALL-E [26] has demonstrated the potential for creativity in generative models, where images are generated based on textual descriptions. Applying similar techniques to ChatGPT could enable the generation of more creative solutions to programming problems. However, fostering creativity in code generation is a complex task, and substantial research is needed to achieve this goal.

4. *Handling incomplete or ambiguous problem statements*: ChatGPT's ability to handle incomplete or ambiguous problem statements is limited by its training data, which predominantly consists of well-structured problem statements [22]. To address this challenge, researchers have explored methods to train GPT models on more diverse and noisy datasets, simulating real-world programming scenarios [27]. Additionally, incorporating interactive components into GPT models can enable them to ask clarifying questions or make reasonable assumptions when faced with ambiguous problem statements [28]. While these approaches hold potential, further research and model refinements are needed to improve ChatGPT's ability to handle incomplete or ambiguous programming problems effectively.

In conclusion, ChatGPT has demonstrated impressive capabilities in natural language understanding and code generation. However, several challenges remain in effectively solving programming problems. By addressing these challenges, future iterations of ChatGPT may be better equipped to tackle complex programming tasks and contribute to the broader field of software development.

*2.3 Codewars*

Established in 2012, Codewars has gained significant recognition as an interactive online platform, enabling developers to enhance their coding skills. This learning is facilitated through code challenges known as "kata" [29] [12]. Crafted by the user community, these katas are ranked according to difficulty, assessed through a "Kata" rating. This rating is derived from votes cast by users who tackled the kata during its beta phase [13]. The katas' complexity spans from 8 kyu (the easiest) to 1 kyu (the most challenging).

Currently, Codewars supports over 20 programming languages, including but not limited to Python, Java, JavaScript, and C++. It also employs a distinctive honor and ranking system to further incentivize user engagement. By providing a highly interactive and competitive learning environment, Codewars fosters a sense of community among its users, thereby transforming coding from an often solitary endeavor into a social experience [30].

Codewars enables users to submit solutions to a kata, which are then automatically evaluated for accuracy against predefined test cases. Users are promptly informed whether their solution is correct.

While there are similar platforms, such as LeetCode [8] or HackerRank [31]. they primarily prepare users for job interviews. In contrast, Codewars encourages users to challenge themselves and elevate their coding proficiency. As such, Codewars' katas span a broader range of difficulties, with the more challenging questions often surpassing those found on other platforms in complexity. This distinctive feature is why we have chosen to utilize katas from Codewars for our research.

III. EXPERIMENT

Initially, we had concerns that GPT might obtain answers from Codewars. Codewars stores all attempts and valid solutions for every kata. Users can view other people's solutions after successfully solving a kata, and the website features the best solutions (based on user judgments and/or internal auto evaluations). To prevent GPT from directly copying answers from Codewars, we attempted to rephrase the questions. However, after a few experiments, we determined that these concerns were unnecessary.

First, the solution styles differ. GPT demonstrates a strong ability to understand natural language and can interpret the requirements correctly. Its solutions appear professional, similar to textbook examples, and follow code conventions precisely; they rarely resemble the best solutions from Codewars. Codewars' top solutions are usually shorter than typical solutions and often consist of a single line of code.

Furthermore, GPT makes mistakes (discussed in more detail later), and even with repeated feedback and multiple rounds of corrections, GPT's solutions can still be incorrect. Thus, we do not worry that GPT might directly copy solutions from Codewars.

*3.4 Simple Questions*

Codewars offers eight different difficulty levels for katas (programming problems), ranging from 8kyu to 1kyu, with 1kyu being the most difficult level. We began our experiment with the easiest level, 8kyu questions.

We posed three 8kyu questions to GPT-3.5 (the default engine during the experiment period) and asked it to provide Python code. The test katas were *Removing element* (takes an array as input, removes every second element), *Get $n^{th}$ even number*, and *You can't code under pressure #1* (write a function to double the input integer).

GPT-3.5 answered two of them correctly in the first test. Only for the second question did it fail to note that we define the first even number as 0, so it took a few attempts to stress this point before it eventually provided the correct answer. The screenshots of dialog between the author and GPT-3.5 are given from Figure 1 to Figure 6.

*Figure 1. GPT-3.5 can answer simple question easily with proper comments.*

*Figure 2. Even give explanation and test cases.*

*Figure 3 After feedback, GPT-3.5 try to improve the code by adding constrains.*

*Figure 4 Provides more test cases, even handle error case.*

*Figure 5. After second round of feedback, it tries to improve, but the code is still wrong. (here we omit the code section to save paper space)*

*Figure 6. After the third time feedback, GPT-3.5 eventually generates the right code.*

We then tested three 7kyu problems: *Max sum between two negatives*, *Page replacement algorithms: FIFO*, and *Hello world - without strings*. GPT-3.5 provided the correct answer for the first 7kyu problem in the second test. In the first test, it missed the requirement of "maximum." However, GPT-3.5 couldn't provide the correct answer for the second kata. It repeatedly gave wrong answers, even with feedback on failed test cases. In contrast, GPT-4 passed this kata. Though it initially failed, it demonstrated clear improvement with feedback on its mistakes and failed test cases, eventually passing all test cases in the fourth test.

## 3.5 Medium Level Questions

After observing the performance of GPT-4 and GPT-3 on simple questions, this study proceeds to evaluate their performance on medium-level questions. To this end, both models were tested on three 6 kyu katas, namely *Convert string to camel case*, *Tortoise racing*, and *Page replacement algorithms: clock*, along with three 5 kyu katas, specifically *Elementary arithmetic – carries count*, *Gigits*, and *Simple fun #273: powerset*.

Although the tasks of converting a string to camel case format and solving a basic arithmetic problem related to tortoise racing might be relatively common, both GPT-3.5 and GPT-4 exhibited proficiency in addressing these challenges. However, GPT-3.5 encountered difficulties with the *Page replacement algorithm: clock*, mirroring its previous struggles with a similar 7 kyu question. Despite receiving multiple feedback attempts, GPT-3.5 demonstrated no discernible improvement. In contrast, GPT-4 successfully answered the question in the second attempt, after being informed that its initial response was incorrect. Notably, GPT-4 managed to rectify the error without being explicitly informed about the problematic test case.

In the case of the 5 kyu katas, GPT-3.5 successfully completed two out of the three assigned tasks after receiving feedback 3 to 5 times. An intriguing observation pertained to the AI's ability to make assumptions. For instance, the Carries count kata required the calculation of carry operations and the subsequent generation of a specific output string. Depending on the number of operations, the output string could adopt one of two forms: "No carry operation" or "n carry operations," where 'n' represents a positive integer. However, both GPT-3.5 and GPT-4 exhibited a propensity to differentiate between singular and plural forms, producing "1 carry operation" as opposed to "1 carry operations" when 'n' equal one. Interestingly, the requirements did not specify this distinction, but the example section of the kata indicated "1 carry operations." This suggests that the AI models disregarded the example section, deriving their output solely from the requirement section.

Despite being classified as a 5 kyu problem, the *Gigits* kata posed a challenge to human programmers due to its convoluted requirements and unconventional implementation method. The following is a description of this kata, as provided by CodeWars:

- *The code consists of four unique digits (from 0 to 9).*
- *Tests will call your solution; you should answer with an array of four digits.*
- *Your input is number of matches (the same digit in the same place) with your previous answer. For the first call input value is -1 (i.e. each new test starts with input -1)*
- *You have to find the code in 16 calls or less. You are the best. Do it.*

*For example*
- *The code is [1, 2, 3, 4]*
- *1st call return [1, 3, 4, 5] will give 1 match in next input*
- *2nd call return [1, 2, 3, 0] will give 3 matches in next input*
- *3rd call return [1, 2, 3, 4] will not give 4 matches in next input, because you're the champion!*

The requirements posed by the Gigits kata can be challenging for most human programmers, who may require considerable time to decipher the task and devise viable solutions. Despite these complexities, GPT-3.5 demonstrated no difficulty in interpreting the requirements and promptly providing solutions. However, these initial solutions contained various errors, and at times, the model failed to produce the correct answer within the stipulated 16 calls. Moreover, GPT-3.5 exhibited no noticeable improvement in its implementations even after receiving feedback.

Conversely, GPT-4 displayed superior performance, generating a valid solution in the third test after receiving only two instances of feedback. Consequently, GPT-4 successfully completed all medium-level katas, showcasing its enhanced problem-solving capabilities in comparison to GPT-3.5.

## 3.6 Advanced Level Questions

We classified 4 kyu and 3 kyu katas as advanced level problems. We asked both GPT-3.5 and GPT-4 to solve three 4 kyu katas—*Greedy thief*, *Binary multiple of 3*, *Counting change combinations*—and three 3 kyu katas—*Total area covered by rectangles*, *Battleship field validator*, and *Simplifying*.

Surprisingly, GPT-3.5 successfully solved two of the three 4 kyu katas on the first attempt: *Binary multiple of 3*, which required writing a regular expression to test if a binary string represented a number divisible by 3 and *Counting change combinations*. Both problems necessitated strong mathematical and programming skills. However, these are well-known problems, leading the authors to suspect that the GPTs might have encountered similar or identical questions during training, enabling GPT-3.5 to answer them correctly in the first round.

*Greedy thief*, an optimization problem, proved more challenging for GPT-3.5. The model struggled with handling items with zero weight, and even after ten rounds of feedback with failed test case information, GPT-3.5 was unable to fix the bugs. In contrast, GPT-4 demonstrated superior programming capability, passing all test cases after only two rounds of feedback.

Although both 3 kyu and 4 kyu katas are classified as advanced level problems, there is a significant gap in difficulty between the two levels. Neither GPT-3.5 nor GPT-4 managed to pass any of the three 3 kyu level katas.

*Total area covered by rectangles* is a relatively easy kata among the three 3 kyu problems, but the real challenge lies in algorithm optimization. Both GPT-3.5 and GPT-4 generated code that passed most simple test cases; however, they failed the complete set of test cases due to timeout issues. Despite repeated feedback, the models were unable to optimize their code, and their performance worsened.

GPT-4 had no difficulty passing the simple test for *Battleship field validator*, while GPT-3.5 struggled. Nonetheless, GPT-4 could not pass the complete set of test cases, and after more than 15 retries with failed test cases as feedback, it showed no observable improvement.

*Simplifying* was perhaps the most challenging kata in our experiment. Neither GPT-3.5 nor GPT-4 could pass even the simple set of test cases for this problem, which required simplifying a formula based on a set of input equations.

*3.7 Hard Questions*

2 kyu and 1 kyu katas are classified as hard questions that can be challenging even for experienced human programmers.

After failed all three kyu katas, we don't expect that current GPTs can solve any 2 kyu and 1 kyu questions. However, to make the experiment more complete, we continue our experiment with three 2 kyu katas: Evaluate mathematical expression, Power tower modulo m, Expression Transpiler, and three 1 kyu katas: Loopover, Game of life reverse, and Express number as sum of four squares.

GPT-3.5 failed all the 6 katas. It did pass a small number of test cases, indicating its capability to interpret the questions properly, but eventually failed due to returning wrong values, raising runtime exceptions, or timeout. And repeating feedback can't help it to improve.

GPT-4 performs better, and it passed Evaluate mathematical expression in the first try, that make us suspect it may have been trained by the same question before. This kata asks to evaluate a mathematical expression input as a string and return a number value. And the solution can't use built in evaluation functions such as eval and exec.

Besides this single exception, GPT-4 also failed other five katas, but it managed to pass about 10% test cases for Loopover contrast to GPT-3.5, that can't pass even one test case of that kata.

*3.8 Summary of the Experiment*

Upon concluding the experiment encompassing eight distinct difficulty levels of katas, the results can be synthesized into the following key observations:

• GPT models exhibited remarkable proficiency in processing and comprehending natural language. They accurately interpreted the requirements of all katas and generated functions that, albeit potentially containing errors, endeavoured to fulfill the intended purpose.

• GPT-4 proved to be considerably more dependable in tackling katas up to 4 kyu. Its solutions successfully passed the entire set of test cases for all examined katas ranging from 8 kyu to 4 kyu.

• Although GPT-3.5 demonstrated the capacity to solve problems up to 4 kyu, its performance exhibited inconsistency.

• Both GPT models could refine their solutions upon receiving feedback. Nevertheless, if they failed to arrive at the correct solution within five rounds of feedback, subsequent feedback did not yield improved accuracy.

• Evidently, GPT models have been trained on certain well-known programming problems, enabling them to correctly address some familiar problem types on their initial attempt.

• At the present stage, it appears that programming tasks at the difficulty level of 3 kyu remain beyond the models' ability to resolve. Despite GPT-4 having passed one 2 kyu kata, it is plausible that this success is attributed to prior training on similar problems.

The comprehensive test results are presented in Table 1. In this table, "Pass" denotes that the GPT model's solution passed all test cases on its first attempt. "Pass*" signifies that the GPT model succeeded in completing the kata with the aid of several rounds of feedback. "Fail" indicates that, even after repeated feedback, the GPT model could not pass the test, and no observable improvements were evident in subsequent attempts.

When a GPT model's attempt fails, we utilize error messages from the CodeWars website as well as failed test cases and their expected results to provide feedback.

*Table 1. The complete test result for GPT-3.5 and GPT-4 on 24 katas*

| # | Kyu | Kata Name | GPT-3.5 | GPT-4 |
|---|---|---|---|---|
| 1 | 8 | Removing elements | Pass | Pass |
| 2 | 8 | Get n$^{th}$ Even Number | Pass* | Pass |
| 3 | 8 | You can't code under Pressure #1 | Pass | Pass |
| 4 | 7 | Max sum between two negatives | Pass* | Pass |
| 5 | 7 | Page replacement algorithms: FIFO | Fail | Pass* |
| 6 | 7 | Hello world - without strings | Pass | Pass* |
| 7 | 6 | Convert string to camel case | Pass* | Pass |
| 8 | 6 | Tortoise racing | Pass | Pass |
| 9 | 6 | Page replacement algorithms: clock | Fail | Pass* |
| 10 | 5 | Elementary arithmetic - carries count | Pass* | Pass |
| 11 | 5 | Gigits | Fail | Pass |
| 12 | 5 | Simple fun #273: powerset | Pass* | Pass |
| 13 | 4 | Greedy thief | Fail | Pass* |
| 14 | 4 | Binary multiple of 3 | Pass | Pass |
| 15 | 4 | Counting change combinations | Pass | Pass |

| 16 | 3 | Total area covered by rectangles | Fail | Fail |
| 17 | 3 | Battleship field validator | Fail | Fail |
| 18 | 3 | Simplifying | Fail | Fail |
| 19 | 2 | Evaluate mathematical expression | Fail | Pass |
| 20 | 2 | Power tower modulo m | Fail | Fail |
| 21 | 2 | Expression Transpiler | Fail | Fail |
| 22 | 1 | Loopover | Fail | Fail |
| 23 | 1 | Game of life in reverse | Fail | Fail |
| 24 | 1 | Express number as sum of four squares | Fail | Fail |

## IV. Discussion

*4.1 Can GPT Replace Human Programmers.*

The remarkable capabilities exhibited by ChatGPT-3 have generated significant impact across various sectors [32], particularly in higher education [33]. This has led to a growing interest in understanding whether GPT can effectively generate programming code, the level of its proficiency, and if it poses a threat to the job security of human programmers.

Despite the limitations of our study, which focuses on programming problems from a single coding website and tests only one programming language, the findings provide preliminary insights to address these questions.

Regarding the first question, can GPT generate programming code? The answer is affirmative, as it has demonstrated the ability to produce reasonable code to solve general programming problems. As for the proficiency level of GPT, based on its performance in solving CodeWars' katas, GPT-4 appears to have reached a level comparable to that of junior programmers.

Our experimental results further indicate that both GPT-3.5 and GPT-4 surpass previous observations on the programming capabilities of GitHub Copilot, a coding suggestion system built on the GPT-3 model [9].

The third and arguably most sensitive question pertains to the potential threat to the job security of human programmers. In the short term, the answer is negative. Although GPT has exhibited the capacity to solve certain programming problems, the tasks demanded of programmers are far more intricate than what current chat-based systems can fully manage. However, considering the rapid advancement of AI, it would not be surprising if more sophisticated systems with comprehensive user interfaces are developed in the future. These systems might then have the potential to replace a considerable number of programming jobs.

*4.2 Strengths and Weaknesses of GPT.*

Although GPT-4 has exhibited impressive programming capabilities, it remains significantly below the level of top human programmers.

The strengths of GPT models include their rapid reading and comprehension capabilities. They can quickly process and "understand" programming questions, providing solutions within minutes. However, a notable weakness is their inability to validate solutions, resulting in repeated errors.

The strengths and weaknesses of GPT models can be attributed to the underlying AI model.

Human programmers employ three distinct mental capabilities to solve coding problems:

1. *Intuition*: This enables programmers to quickly devise solutions once they understand the problem, assuming it is not overly complex. This capability is developed through learning and practice. GPT models employ deep neural networks to simulate this capability [16]. Although deep neural networks can produce relatively accurate outcomes for various inputs, there is no guarantee of solution quality.

2. *Self-validation*: After devising an initial solution, programmers can write code and validate it through test cases, refining or discarding their approach based on the results. AI can use tree search algorithms for validation [14], but this may demand significant computational power as the search space can expand rapidly.

A combination of these approaches can yield optimized results, as exemplified by AlphaGo [32]. AlphaGo's algorithm, which combines deep neural networks and Monte Carlo Tree Search (MCTS), allowed it to outperform traditional Go programs and achieve unprecedented success [32] [34]. MCTS, a renowned reinforcement learning algorithm, could potentially be integrated into the GPT training process. However, during the chat process, the need for rapid responses might restrict the application of more time-consuming algorithms like MCTS.

3. *Creative thinking*: Unlike Go, where the number of valid moves is finite, there is no fixed list of algorithms to solve a programming problem. Although creative thinking is crucial for top programmers, no AI approach currently emulates this human capability, presenting an intriguing avenue for future research.

GPT models are trained using deep neural networks, equipping them with the first mental capability of programmers but lacking the subsequent two. In particular, for the third mental capability even though there are some researches in this direction [26] [15], but there is no mature approach and the results are still far lower than the capabilities observed in top humans.

## V. Conclusion and Future Work

This paper presents a novel evaluation of the programming capabilities of GPT models on coding problems of varying difficulty levels. Our experiment, involving GPT-3.5 and GPT-4, demonstrates a clear boundary at the 3kyu difficulty level. Problems from 1kyu to 3kyu remain unsolved, suggesting that there's still substantial room for improvement in GPT models' problem-solving abilities.

We've identified that for GPT models to better mimic human intelligence in problem-solving, it is crucial for these models to acquire validation and creative thinking capabilities. Current models' performance, while impressive in solving problems up to a certain difficulty level, falls short in more complex problems requiring a higher level of creativity and validation.

This study has introduced a measure of coding problem complexity that distinguishes between the capabilities of human programmers and AI models like GPT. We've proposed that problem complexity should be measured not only by the difficulty of the problem but also by the time taken to solve it. This measurement reflects more accurately the nature of problem-solving in coding and offers a new direction for evaluating the capabilities of AI models in programming.

Looking forward, it's critical to enhance the training methods and the design of AI models to incorporate the validation and creative thinking capabilities. Such advancements can help AI models break the identified boundary and perform at the highest human programming level.

Additionally, we believe that more research should be devoted to establishing an objective measurement of programming problem difficulty. Such a measurement can provide a more accurate benchmark for evaluating AI programming capabilities, thereby furthering the development of more competent and intelligent AI models.

Our future work will involve refining the proposed measure of problem complexity and applying it to a broader range of problems and AI models. Moreover, we plan to investigate techniques for enhancing the validation and creative thinking capabilities of AI models, aiming to break the identified boundary and bring these models closer to human-level programming prowess.

## Bibliography

[1] OpenAI, "ChatGPT," OpenAI, [Online]. Available: https://chat.openai.com. [Accessed 30 05 2023].

[2] A. Kashefi and T. Mukerji, "ChatGPT for Programming Numerical Methods," arXiv, 2023.

[3] N. M. S. Surameery and M. Y. Shakor, "Use Chat GPT to Solve Programming Bugs," *IJITC,* vol. 3, no. 1, 2023.

[4] R. Poldrack, T. Lu and G. Beguš, "AI-assisted coding: Experiments with GPT-4," arXiv, 2023.

[5] A. Cheshkov, P. Zadorozhny and R. Levichev, "Technical Report: Evaluation of ChatGPT Model for Vulnerability Detection," arXiv, 2023.

[6] J. Savelka, A. Agarwal, C. Bogart and M. Sakr, "Large Language Models (GPT) Struggle to Answer Multiple-Choice Questions about Code," arXiv, 2023.

[7] J. Savelka, A. Agrarwal, C. Bogart, Y. Song and M. Sakr, "Can Generative Pre-trained Transformers (GPT) Pass Assessments in Higher Education Programming Courses?," arXiv, 2023.

[8] LeetCode, "LeetCode," [Online]. Available: https://leetcode.com/. [Accessed 12 05 2023].

[9] N. Nguyen and S. Nadi, "An Empirical Evaluation of GitHub Copilot's Code Suggestions," in *IEEE/ACM 19th International Conference on Mining Software Repositories (MSR)*, 2022.

[10] HackerRank, "HackerRank," [Online]. Available: https://www.hackerrank.com/. [Accessed 30 05 2023].

[11] A. Karmakar, J. Prenner and R. Robbes, "Codex Hacks HackerRank: Memorization Issues and a Framework for Code Synthesis Evaluation," ArXiv, 2022.

[12] CodeWars, "What is Kata," [Online]. Available: https://docs.codewars.com/concepts/kata/. [Accessed 12 05 2023].

[13] Codewars, "Reviewing a Kata," [Online]. Available: https://docs.codewars.com/curation/kata/. [Accessed 12 05 2023].

[14] C. B. Browne, E. Powley, D. Whitehouse and S. M. Lucas, "A survey of Monte Carlo tree search methods," *IEEE Transactions on Computational Intelligence and AI in Games,* vol. 4, no. 1, pp. 1-43, 2012.

[15] S. Colton, R. Mántaras and O. Stock, "Computational Creativity: Coming of Age," *AI Mag.,* vol. 30, pp. 11-14, 2009.

[16] Y. LeCun, Y. Bengio and G. Hinton, "Deep learning," *Nature,* vol. 521, no. 7553, pp. 436-444.

[17] D. Amodei, C. Olah, J. Steinhardt, P. Christiano, J. Schulman and D. Mané, *Concrete Problems in AI Safety,* 2016.

[18] T. Brown, B. Mann, N. Ryder, M. Subbiah, J. Kaplan and P. Dhariwal, "Language Models are Few-Shot Learners," in *Advances in Neural Information Processing Systems*, 2020, pp. 1877--1901.

[19] T. Besold, J. Hernández-Orallo and U. Schmid, "Can Machine Intelligence be Measured in the Same Way as Human intelligence?," in *Künstl Intell*, 2015.

[20] J. P. Adams and S. Turner, "Problem Solving and Creativity for Undergraduate Engineers: process or product?," *Innovation, Good Practice and Research in Engineering Education,* 2008.

[21] L. Hancox-Li, "Robustness in machine learning explanations: Does it matter?," in *Proceedings of the 2020 conference on fairness, accountability, and transparency*, 2020.

[22] A. Radford, K. Narasimhan, T. Salimans and I. Sutskever, "Improving language understanding by generative pre-training," OpenAI, 2018.

[23] G. Lample and F. Charton, *Deep Learning for Symbolic Mathematics,* arXiv, 2019.

[24] M. Allamanis, E. Barr and P. Devanbu, "A Survey of Machine Learning for Big Code and Naturalness," *ACM Computing Surveys,* vol. 51, no. 4, pp. 1-37, 2018.

[25] S. Gulwani, "Programming by Examples: Applications, Algorithms, and Ambiguity Resolution," in *International Joint Conference on Automated Reasoning*, 2016.

[26] A. Ramesh, M. Pavlov and G. Goh, "Zero-Shot Text-to-Image Generation," in *Proceedings of Machine Learning Research*, 2021.

[27] R. Gupta, S. Pal, A. Kanade and S. Shevade, "DeepFix: Fixing Common C Language Errors by Deep Learning," in *Proceedings of the AAAI Conference on Artificial Intelligence*, 2017.

[28] S. Amershi and D. Weld, "Guidelines for Human-AI Interaction," in *Proceedings of Chi 2019*, 2019.

[29] Bustle, "6 Best Resources To Learn How To Code," [Online]. Available: https://www.bustle.com/p/6-best-resources-to-learn-how-to-code-61104. [Accessed 12 05 2023].

[30] H. Jens, "Teach coding with games: a review of Codewars and CodeCombat," Opensource.com, 2015.

[31] HackerRank, "HanckerRank," [Online]. Available: https://www.hackerrank.com/. [Accessed 12 05 2023].

[32] S. A. George and H. A. S. George, "A Review of ChatGPT AI's Impact on Several Business Sectors," *Partners Universal International Innovation Journal,* vol. 1, no. 1, pp. 9-23, 2023.

[33] K. Malinka, M. Perešíni, A. Firc, O. Hujňák and F. Januš, *On the Educational Impact of ChatGPT: Is Artificial Intelligence Ready to Obtain a University Degree?,* malinka2023educational, 2023.

[34] DeepMind, "AlphaGo," [Online]. Available: https://www.deepmind.com/research/highlighted-research/alphago.

[35] D. Silver, A. Huang, C. J. Maddison, A. Guez, L. Sifre, G. van den Driessche, J. Schrittwieser, I. Antonoglou and V. Panneershelvam, "Mastering the game of Go with deep neural networks and tree search," *Nature,* vol. 529, no. 7587, 2016.